\documentclass[10pt]{article} 
\usepackage[accepted]{rlc}

\usepackage{amssymb}            
\usepackage{mathtools}          
\usepackage{mathrsfs}           
\mathtoolsset{showonlyrefs}     
\usepackage{graphicx}           
\usepackage{subcaption}         
\usepackage[space]{grffile}     
\usepackage{url}                

\usepackage[utf8]{inputenc} 
\usepackage[T1]{fontenc}    
\usepackage{hyperref}       
\usepackage{url}            
\usepackage{booktabs}       
\usepackage{amsfonts}       
\usepackage{nicefrac}       
\usepackage{microtype}      
\usepackage{xcolor}         

\usepackage{multirow}
\usepackage{booktabs}
\usepackage{graphicx}
\usepackage{wrapfig}
\usepackage{enumitem}

\title{D5RL: Diverse Datasets for Data-Driven Deep Reinforcement Learning}

\author{%
  Rafael Rafailov$^{1}$\thanks{Denotes equal contribution}, Kyle Hatch$^{1*}$, Anikait Singh$^2$, Laura Smith$^2$, Aviral Kumar$^2$, \\ \textbf{Ilya Kostrikov$^2$, Philippe Hansen-Estruch$^2$, Victor Kolev$^1$, Philip Ball$^2$,} \\ \textbf{Jiajun Wu$^1$, Chelsea Finn$^1$, Sergey Levine$^2$}\\[1em] 
  $^1$Stanford University, $^2$UC Berkeley \\
}

\begin{document}

\maketitle

\begin{abstract}
Offline reinforcement learning algorithms hold the promise of enabling data-driven RL methods that do not require costly or dangerous real-world exploration and benefit from large pre-collected datasets. This in turn can facilitate real-world applications, as well as a more standardized approach to RL research. Furthermore, offline RL methods can provide effective initializations for online finetuning to overcome challenges with exploration. However, evaluating progress on offline RL algorithms requires effective and challenging benchmarks that capture properties of real-world tasks, provide a range of task difficulties, and cover a range of challenges both in terms of the parameters of the domain (e.g., length of the horizon, sparsity of rewards) and the parameters of the data (e.g., narrow demonstration data or broad exploratory data). While considerable progress in offline RL in recent years has been enabled by simpler benchmark tasks, the most widely used datasets are increasingly saturating in performance and may fail to reflect properties of realistic tasks. We propose a new benchmark for offline RL that focuses on realistic simulations of robotic manipulation and locomotion environments, based on models of real-world robotic systems, and comprising a variety of data sources, including scripted data, play-style data collected by human teleoperators, and other data sources. Our proposed benchmark covers state-based and image-based domains, and supports both offline RL and online fine-tuning evaluation, with some of the tasks specifically designed to require both pre-training and fine-tuning. We hope that our proposed benchmark will facilitate further progress on both offline RL and fine-tuning algorithms. Website with code, examples, tasks, and data is available at \url{https://sites.google.com/view/d5rl/}
\end{abstract}

\section{Introduction}
\vspace{-0.25cm}


Offline reinforcement learning algorithms hold the promise of enabling data-driven RL methods that do not require costly or dangerous real-world exploration, and benefit from pre-collected datasets~\citep{levine2020offline,gulcehre2020rl,agarwal2020optimistic}. The latter especially is of significant relevance in the modern age of data-driven machine learning, where training on large datasets has repeatedly been shown to be a critical ingredient for effective generalization~\citep{lecun2015deep,krizhevsky2017imagenet} and even emergent capabilities~\citep{wei2022emergent}. Furthermore, offline RL methods can provide effective initializations for online finetuning, overcoming challenges with exploration and providing an effective formula for fast online training suitable for the real world. However, while supervised learning methods that operate on large pre-collected datasets can effectively evaluate on test sets sampled from real-world data, offline RL algorithms that train on data must still be \emph{validated} through online interaction to measure their effectiveness, even if no online interaction is required during training. Therefore, evaluating progress on offline RL methods requires effective and challenging benchmarks that can provide for accessible evaluation in simulation, while still providing a degree of realism in terms of reflecting the properties of real-world systems, and covering a range of challenges both in terms of the parameters of the domain (e.g., length of the horizon, sparsity of rewards) and the parameters of the data (e.g., narrow demonstration data or broad exploratory data). While considerable progress in offline RL in recent years has been enabled by simpler benchmark tasks, the most widely used datasets are increasingly saturating in performance~\citep{fu2020d4rl,gulcehre2020rl}, might fail to reflect properties of realistic tasks, and might not cover some of the most significant use cases, such as online finetuning from offline initialization.
\begin{figure}
    \centering
    \includegraphics[width=.92\textwidth]{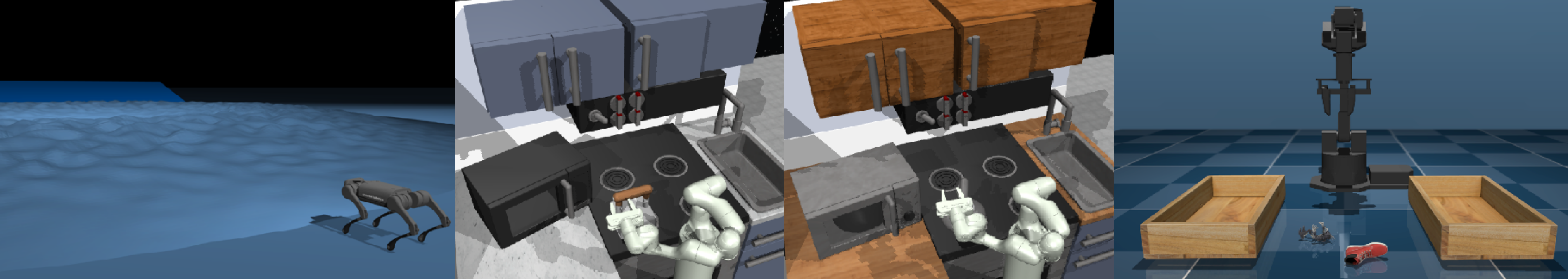}
    \caption{\footnotesize A visualization of the environments in our proposed benchmark. We provide datasets for training locomotion policies for the A1 robot (left), learning manipulation in randomized vision-based kitchen-like environments with a Franka robotic arm (middle), and learning multi-stage pick-and-place tasks with a WidowX low-cost robotic manipulator (right). Each domain is accompanied by several datasets with different properties and evaluates a distinct aspect of offline RL and offline training with online finetuning.}
    \label{fig:teaser}
    \vspace{-.5cm}
\end{figure}
In this paper, we propose a new benchmark for offline RL that focuses on realistic simulations of robotic manipulation and locomotion environments, based on models of real-world robotic systems, and comprising a variety of data sources, including scripted data, play-style data collected by human teleoperators, and other data sources. Our proposed benchmark covers state-based and image-based domains, and supports both offline RL evaluation and evaluation with online finetuning, with some of the tasks specifically designed to require both pre-training and finetuning. We hope that our proposed benchmark will facilitate further progress on both offline RL algorithms and algorithms designed for online finetuning from offline initialization.


We present an overview of the environments in our benchmark in Figure~\ref{fig:teaser}, which include realistic simulated models of real-world robotic platforms, such as the A1 quadruped and the Franka robotic arm. Aside from providing a more challenging and up-to-date range of tasks and datasets compared to prior work~\citep{fu2020d4rl,gulcehre2020rl}, our tasks cover a range of factors that are either rarely covered in prior benchmarks, or rarely appear in combination. The A1 tasks specifically evaluate online finetuning: these tasks are designed such that offline initialization should provide for basic but low-performance capability (e.g., not falling), while online finetuning is required for maximally effective gaits. The visual Franka kitchen environments evaluate visual perception, environment variability (accomplished via randomization), and ability to use ``play-style'' diverse data collected by real humans via teleoperation. The visual WidowX pick-and-place environments evaluate the ability to ``stitch together'' distinct phases of manipulation skills to accomplish multi-stage behaviors. While prior datasets evaluate stitching (e.g., the AntMaze task in D4RL~\citep{fu2020d4rl}), it is rarely evaluated in combination with visual perception in widely accepted benchmarks.

We provide a comprehensive description of our proposed tasks and corresponding datasets, as well as high-quality implementations of a number of widely used offline RL and online finetuning methods that we evaluate on our benchmark. We show that current RL methods can and do perform well on lower dimensional tasks such as locomotion, but do not reliably scale to realistic robot scenarios. In our experiments all current RL algorithms do not reliably out-perform simple BC approaches on realistic robot tasks that involve more complex motions, chaining together multiple objectives (“stitching”) or realistic distribution shifts. In fact, they do not demonstrate any of these qualities, i.e. effective dynamic programming or robustness. We believe this shows that current RL approaches are not suitable to large-scale realistic robotic scenarios, where simple imitation learning with large capacity models (such as the RT-* systems) is used. We have structured our tasks and datasets to provide a testing bed for new algorithms for such potential applications, while still maintaining simplicity, speed and accessibility (all our experiments take less than 12 hours on a single GPU) as compared to larger scale benchmarks. We hope that this will provide a solid foundation for future progress on both offline reinforcement learning and online finetuning from offline initialization.

\vspace{-0.3cm}
\section{Related Work}
\vspace{-0.3cm}

Benchmarking in reinforcement learning been a persistent challenge, with effective benchmarks needing to balance accessibility (i.e., tasks that are feasible to address this current methods and not too onerous computationally) with the desired for broad coverage of task properties and a high degree of realism and complexity~\citep{duan2016benchmarking,brockman2016openai,wu2017flow,wang2019benchmarking,hubbs2020or,yu2020meta}. Striking this balance is arguably a greater challenge in RL than in other fields. First, RL algorithms can be applied to a wide range of tasks with very different properties, including varying time horizons, levels of reward sparsity, dimensionality, and other ingredients~\citep{osband2019behaviour}. Second, RL algorithms can be computationally very demanding, requiring long training runs that make it difficult to include large numbers of very complex tasks in every evaluation~\citep{henderson2018deep,agarwal2021deep}. Third, the capabilities of RL methods have advanced significantly over the past decade, and benchmarks can quickly become saturated, necessitating more complex tasks to be added~\citep{dulac2021challenges}. This makes designing a good benchmark in RL a major challenge. Our work focuses specifically on benchmarking offline RL methods, and aims to strike a balance between covering task complexity and a variety of task ingredients with providing a convenient simulated evaluation protocol and a mixture of image-based and state-based tasks.

In recent years, a number of benchmarks have been proposed for offline RL, though such benchmarks typically have a number of shortcomings that have proven difficult to fully alleviate while balancing the aforementioned challenges. Early work on deep offline RL focused either on customized evaluations without proposing standard benchmarks~\citep{vecerik2017leveraging,hester2018deep,kalashnikov2018qt}, or else proposed simple benchmark tasks that utilized replay buffers from successful RL runs~\citep{fujimoto2019off,kumar2019stabilizing,agarwal2020optimistic}. The latter generally does not evaluate the performance of offline RL methods effectively, as realistic data might be highly sub-optimal and might require ``stitching'' together parts of different sub-optimal trajectories to create ones that are more optimal -- a property rarely captured by data collected by fully or partially trained RL policies themselves~\citep{fu2020d4rl,levine2020offline}. Several more recent offline RL benchmarks have sought to include more realistic data distributions, more complex tasks (including vision-based tasks), and other ingredients that are intended to more accurately represent realistic offline RL problems~\citep{gulcehre2020rl,liu2022finrl,kurenkov2022showing,kuo2022health,qin2022neorl,lu2022challenges}. Some works have proposed protocols for benchmarking offline pretraining with online finetuning~\citep{kostrikov2021offline,nair2020awac,song2022hybrid,nakamoto2023calql}, though this has not been rigorously systematized in prior work. Perhaps the most widely used benchmark suite in offline RL today is D4RL~\citep{fu2020d4rl}. However, the D4RL tasks are increasingly saturated in performance, and many of the tasks do not effectively reflect the challenges of realistic offline RL tasks: the MuJoCo locomotion tasks in D4RL are still largely based on RL replay buffers, and the more complex ``maze'' tasks, which do feature sub-optimal data and require stitching or recombining parts of the sub-optimal trajectories, are limited in difficulty and variety. Our benchmark aims to address these limitations in several ways. We focus specifically on robotics-themed tasks -- although RL can address a far greater range of problems, we believe that this focus is reasonable for providing a balance between specificity (i.e., not so much breadth that no single method can address all tasks) and coverage (i.e., still capturing different challenges in RL). Within this theme, our tasks all reflect realistic simulated models of robotic systems based on actual robot URDF specifications, in contrast to D4RL, which uses simple ``fictional'' rigid body systems. Our tasks include both state-based and image-based tasks, both sparse and dense rewards, and multi-stage tasks. Additionally, we propose tasks suitable for offline pre-training with online finetuning, something that has not been rigorously formalized in current widely used benchmarks.

Offline RL algorithms themselves have made significant progress in recent years as well~\citep{fujimoto2019off,kumar2019stabilizing,kumar2020conservative,agarwal2020optimistic,kostrikov2021offline,nair2020awac,song2022hybrid,cheng2022adversarially,nakamoto2023calql}. A full survey of all recent research on offline RL is outside the scope of this paper, but we do make an attempt to benchmark representative examples of some of the widely used algorithm classes, including pessimistic or conservative algorithms~\citep{kumar2020conservative,nakamoto2023calql}, algorithms based on implicit backups~\citep{kostrikov2021offline}, algorithms based on behavioral cloning regularization~\citep{fujimoto2021minimalist}, algorithms that utilize diffusion models~\citep{hansenestruch2023idql}, and methods designed specifically for efficient online training by leveraging offline data~\citep{ball2023rlpd}. We hope that by proposing a new benchmark that addresses the limitations of prior datasets and environments we will provide a more effective means for algorithms researchers to make further advances in the future.

\section{Preliminaries and Background}

Reinforcement learning is formalized through the concept of Markov Decision Process (MDP) $\mathcal{M} = (\mathcal{S}, \mathcal{A}, P, R, \rho, \mathcal{\gamma})$, where $\mathcal{S}$ is the state space, $\mathcal{A}$ is the action space, $P(s'|s, a)$ is the transition probability, $R(s, a)$ is the reward function, $\rho$ is the initial state distribution and $\gamma$ is a discount factor. The goal of reinforcement learning is to find a policy $\pi(a|s)$ that maximizes the expected reward:
\begin{equation}
    J(\pi) = \mathbb{E}_{\rho, P, \pi}\left[\sum_{t=0}^{\infty}\gamma^tR(s_t, a_t)\right]
\end{equation}
In the standard RL setting the policy is given access to the MDP and can sample trajectories to collect additional data. On-policy algorithms iterate between data collection and policy updates, and discard the collected data after each update, which makes them sample inefficient. Off-policy algorithms collect data in a replay buffer, which is then repeatedly used to update the policy. 

\textit{Offline reinforcement learning} also reuses previously collected data, but unlike off-policy algorithms it does not have access to the MDP during training and only utilizes a static dataset. These algorithms need to be able to handle distribution shift between their training datasets and deployment. Moreover they  need to be able to utilize a variety of data sources and qualities, such as prior training runs, deployments, data from different agents or human-generated data.

Additionally, prior offline data can be leveraged with online RL, either by \emph{pre-training} offline and \emph{finetuning} online~\citep{nair2020awac,kostrikov2021offline}, or by training online but including the prior data in a replay buffer (i.e., joint offline and online training)~\citep{song2022hybrid,ball2023rlpd}. The challenge in this setting is for the policy to effectively utilize the offline data to reach high performance in a sample-efficient way. 

Our proposed tasks and datasets can be used for both problems, pure offline RL and offline-to-online fine-tuning, and we evaluate both settings in our experiments.

\vspace{-0.3cm}
\section{Challenges in Offline RL Evaluation}
\vspace{-0.3cm}

Our benchmark environments and datasets aim to cover a range of challenges that are likely to be encountered by offline RL algorithms aiming to learn effective policies for real-world tasks. Some of these challenges, like temporal compositionality (``stitching''), have been addressed via simpler and less realistic environments in prior benchmarks~\citep{fu2020d4rl}. Other challenges, like the use of visual observations, are present in prior tasks~\citep{gulcehre2020rl}, but in combination with less realistic data distributions, such as data from the replay buffer of online RL runs. We discuss some of these challenges below, and in Section~\ref{sec:tasks} discuss how our tasks instantiated some of these challenges.

\textbf{Diverse and realistic robot systems:} We evaluate simulated environments based on the A1 legged robot, the WidowX low-cost manipulation platform, and the Franka Emika robot arm. All of the robots are based on their actual URDFs (definitions of robot morphology), controlled in ways that are analogous to their real-world counterparts (e.g., position control or end-effector control).

\textbf{Realistic observation spaces:} Previous offline RL benchmarks, such as \cite{fu2020d4rl}, mostly focus on low-dimensional state observations, even for more complex robotic tasks. In real scenarios, ground-truth system states are not available; correspondingly, our tasks and datasets use high-dimensional and multi-view RGB camera observations as well as robot proprioception. This creates additional challenges, such as partial observations and state estimation, so the reinforcement learning agent needs to learn robust representations in addition to behaviors.

\textbf{Generalization to environment variability:} One of the central challenges for real-world embodied systems is dealing with variability in the environment, from simple variation in appearance to changes in object pose. In real settings, even small changes in the environment can significantly affect agent performance. There are a number of realistic environments and simulators specifically designed to evaluate an agent's robustness to visual conditions, such as \cite{dosovitskiy2017carla} in autonomous driving or \cite{szot2021habitat, xia2019gibson}. However, these are very heavy-weight in terms of software, datasets and compute requirements, which makes them hard for wide adoption and fast algorithm iteration. To strike a balance between realistic challenges, simplicity, and accessibility, we use comparatively more lightweight MuJoCo-based simulations, but with significantly more realistic visuals and variability. To evaluate agent's robustness and ability to generalize, some of our tasks vary the objects the robot needs to manipulate and randomize their arrangement. In addition, on the observation side we introduce a number of distractors by varying textures, object colors, lighting conditions and camera angles. 

\textbf{Datasets:} We aim to explicitly evaluate datasets used in realistic robot applications that present challenges for current algorithms. Towards that goal we focus on narrow data distributions from scripted planners as well as human-generated data. While some prior benchmarks also include scripted and human-generated data~\citep{fu2020d4rl}, many of the previously studied tasks consist of replay buffers from online RL runs~\citep{gulcehre2020rl}, which may not be reflective of the data distributions on which we might want to train real-world systems. In the WidowX platform we generate object manipulation data using (sub-optimal) scripted planners. In the Franka domain, we collected 20 hours of new human teleoperation data, and also include tasks based on the datasets from prior work~\citep{gupta2019relay,fu2020d4rl}, but rendered out with visual observations rather than low-dimensional state.
We include both expert-level demonstrations from an experienced teleoperator, as well as play data from several teleoperators with different levels of experience. We believe these data distributions are realistic and provide significant challenges to current algorithms, since they are not multi-task, have significant multi-modality with various levels of quality, and are not executed in any particular order, which introduces significant challenges for dynamic programming algorithms.

\textbf{Temporal compositionality and multi-stage tasks:} One of the most appealing properties of offline RL methods is the ability to combine parts of sub-optimal behaviors and compose them into new behaviors that complete more complex tasks more effectively~\citep{levine2020offline,fu2020d4rl}. One of the ways that offline RL can do this is by exploiting temporal compositionality: if the algorithm understands that it's possible to reach C from B, and to reach B from A, then it should be able to figure out how to reach C from A. This can enable solving multi-stage tasks (such as sorting multiple objects) by composing shorter-horizon primitive behaviors. Our benchmarks are designed to evaluate temporal compositionality both by composing task-agnostic or multi-task sub-optimal data (e.g., ``play'' data) into longer and more optimal tasks, and by composing single-step behaviors to solve multi-stage tasks, such as sorting objects.

\textbf{Online training from offline data:} In many cases, we might want to use offline RL not to acquire a policy that we deploy in the real world in zero shot, but rather to provide an initialization for online training for a skill that would be difficult (or dangerous) to acquire entirely from scratch. This can be done either via offline pre-training and finetuning~\cite{nair2020awac,kostrikov2021offline,nakamoto2023calql}, or by using online RL algorithms that can incorporate offline data~\citep{song2022hybrid,ball2023rlpd}. Prior benchmarks rarely evaluate this setting, and prior works studying this setting tend to use a non-standard combination of tasks adopted by the community.


\vspace{-0.25cm}
\section{D5RL: Diverse Datasets for Data-Driven Deep Reinforcement Learning}
\label{sec:tasks}
\vspace{-0.25cm}

In this section, we describe the individual tasks in our benchmark, and relate them to the challenges outlined in the preceding section. Each of our tasks reflects a realistic simulated model of a robotic system, using the URDF of the corresponding robot and a simulated environment to enable plausible interactions. Although our goal is primarily to enable rapid algorithms development rather than to provide a framework for robotics research, we believe that this added degree of realism increases the chances that algorithmic developments made with our benchmark will translate into good real-world performance. Beyond the below descriptions, additional details about the environments and datasets are provided in Appendix A.

\vspace{-0.25cm}
\subsection{Legged Locomotion}
\vspace{-0.25cm}

\begin{wrapfigure}{r}{0.3\textwidth}
\vspace{-0.55cm}
  \begin{center}
    \includegraphics[width=0.27\textwidth]{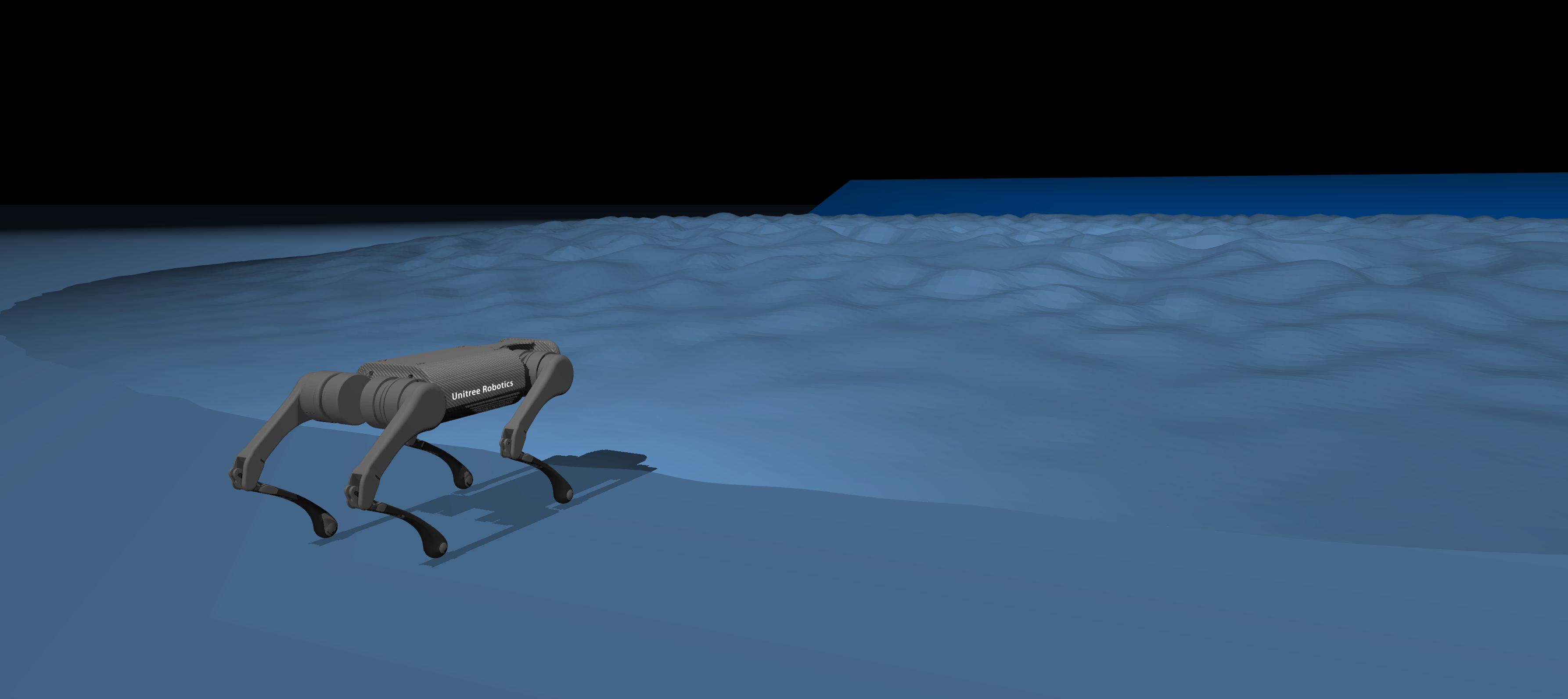}
  \end{center}
  \vspace{-0.35cm}
  \caption{\footnotesize{\textbf{Hiking task}. The A1 robot at the start of the course in front of a randomized terrain.}}
  \label{fig:a1-hiking}
  \vspace{-0.5cm}
\end{wrapfigure}

The goal of the legged locomotion tasks is to study the efficacy of offline RL methods in handling low-level control problems with complex dynamics. We set up these tasks on a simulated Unitree A1 robot platform and require learning policies from low-dimensional proprioceptive observations and do not require visual perception. Additionally, in our experiments, we also evaluate both the offline training performance and performance after online fine-tuning for these tasks. Concretely, we construct three offline datasets, each of which aim to learn different types of locomotion skills as follows: 

\textit{Interpolate Speed:} The goal is to control the A1 at a particular speed level, within the range of speeds that were observed in the training data. For this, we first collect a dataset by training an A1 to track 3 speeds: 0.5, 0.8, and 1.0 m/s, containing experience from the agents' initial exploration to expert-level performance on those tasks, and the goal is to adapt to a speed value of 0.75 m/s, that lies within the range of speeds observed in the dataset. To compute rewards for offline RL training, we label each transition with how accurately it tracks the target speed of 0.75 m/s. 

\textit{Extrapolate Speed:} Using the same dataset as the Interpolate Speed task, this task instead tests the ability of an algorithm to be able to acquire a policy that can run at a higher speed of 1.25 m/s. This task presents a challenge for offline RL methods as the optimal policy that runs at the higher speed lies outside the support of the offline dataset, which means that this task presents a significant room for improvement with online fine-tuning.

\textit{Hiking:} Finally, we construct a task that aims to test the efficacy of offline RL at learning policies when interacting with the complex dynamics induced by navigating on a hiking course (shown in Fig.~\ref{fig:a1-hiking}). This task still utilizes a offline dataset that depicts navigation on a flat terrain, but is distinct in that the policy is deployed on a hiking course, and not a flat terrain. Our hiking course presents varied terrains consisting of randomly generated rolling bumps as well as inclines and declines for evaluation and our goal is to navigate the policy to the center of the course without falling.

\vspace{-0.25cm}
\subsection{Franka Kitchen Manipulation Environment}
\vspace{-0.25cm}

The goal of this environment is to study offline RL and online fine-tuning from realistic but sub-optimal human-generated data, evaluate settings with variability in the appearance and placement of objects to measure generalization, and handle multiple visual observations. Near-optimal and sub-optimal human-collected data, which can run the gamut from demonstrations to unstructured ``play'', represents a realistic source of training data for offline RL, which has been studied in several prior works~\citep{lynch2020learning,gupta2019relay,mandlekar2021matters}. Additionally, generalization over object placement and appearance is very important in real-world settings, but is rarely evaluated in RL benchmarks~\citep{cobbe2019quantifying,cobbe2020leveraging}. Therefore, we hope that this task will cover a range of challenges that are underrepresented in prior work. This environment consists of a Franka Emika robot in a simulated kitchen setting, and data is collected via VR-based tele-operation by real people. We introduce several environments that pose different challenges for current data-centric RL algorithms.

\vspace{-0.25cm}
\subsubsection{Standard Franka Kitchen Environment}
\label{sec:franka_standard}
\vspace{-0.25cm}
\begin{wrapfigure}{r}{0.4\textwidth}
\vspace{-0.85cm}
  \begin{center}
    \includegraphics[width=0.185\textwidth]{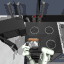}
    \includegraphics[width=0.185\textwidth]{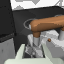}
  \end{center}
  \vspace{-0.35cm}
  \caption{\footnotesize{Observations for the Standard Franka Kitchen tasks consist of two $64\times 64$ RGB images from an a top-down and a wrist camera, as well as robot proprioception.}}
  \label{fig:franka_standard}
  \vspace{-0.35cm}
\end{wrapfigure}

For an easier starting point, we adapt the Franka Kitchen environment which was introduced by \citet{gupta2019relay} and was also part of the D4RL \citep{fu2020d4rl} benchmark. The objective in this environment is to manipulate a set of 4 pre-specified objects. We modify the task to utilize multiple image observations rather than ground truth object locations, thus providing an observation space that more realistically reflects robotic manipulation scenarios. The agent receives a sparse reward of +1 for every object manipulated into the correct configuration.

\textbf{Datasets:} We use the same datasets as \citep{lynch2019play, fu2020d4rl}, which consists of expert-level demonstrations for different combinations of four objects, executed in a fixed order. In total there are 513 total trajectories of varying length split across 22 task combinations. Our observation space consists of two $64\times64$ images from a side-view and wrist cameras \cite{hsu2022seeing} as shown in Fig. \ref{fig:franka_standard}, as well as robot proprioception. 

\textbf{Tasks}: We consider two settings, similar to \cite{fu2020d4rl}:
\begin{enumerate}[leftmargin=*]

\item\textbf{Mixed:} In this environment the agent needs to rearrange the microwave, kettle, light switch and slide door objects, and there are several expert demonstrations in the offline dataset for that combination of objects.

\item\textbf{Partial:} In this setting the agent needs to manipulate the microwave, kettle, bottom burner knob and light switch objects, which are never encountered together in any of the trajectories in the offline dataset. This requires the agent to learn combinatorial generalization capabilities. We note that this is different from the dynamic programming or "stitching" problem, since there is no sequence of states in the dataset that reach the optimal solution.
\end{enumerate}


\vspace{-0.25cm}
\subsubsection{Randomized Franka Kitchen Environment}
\label{subsection:randomized_kitchen}
\vspace{-0.2cm}
\begin{figure}
    \centering
    \includegraphics[width = 0.105\textwidth]{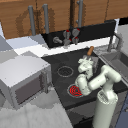}
    \includegraphics[width = 0.105\textwidth]{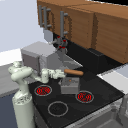}
    \includegraphics[width = 0.105\textwidth]{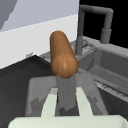}
    \includegraphics[width = 0.105\textwidth]{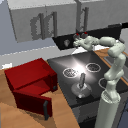}
    \includegraphics[width = 0.105\textwidth]{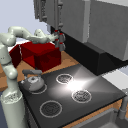}
    \includegraphics[width = 0.105\textwidth]{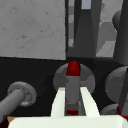}
    \includegraphics[width = 0.105\textwidth]{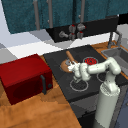}
    \includegraphics[width = 0.105\textwidth]{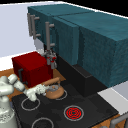}
    \includegraphics[width = 0.105\textwidth]{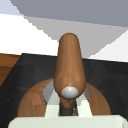}

    \includegraphics[width = 0.105\textwidth]{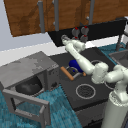}
    \includegraphics[width = 0.105\textwidth]{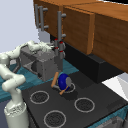}
    \includegraphics[width = 0.105\textwidth]{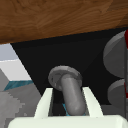}
    \includegraphics[width = 0.105\textwidth]{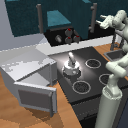}
    \includegraphics[width = 0.105\textwidth]{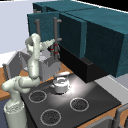}
    \includegraphics[width = 0.105\textwidth]{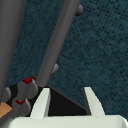} 
    \includegraphics[width = 0.105\textwidth]{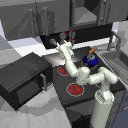}
    \includegraphics[width = 0.105\textwidth]{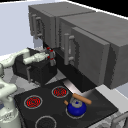}
    \includegraphics[width = 0.105\textwidth]{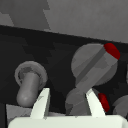}   
    \caption{\small Observations from the Randomized Kitchen environment consist of two $128\times 128$ RGB images from side-cameras, $128\times128$ RGB image from a wrist camera, and robot proprioception. The environment includes several different types of kettles and microwaves, which require different grasps. Moreover, their locations are randomized across the scene. Textures, lighting conditions, and camera angles are also varied across episodes.}
    \label{fig:domain_shift}
\end{figure}

We include a version of the Franka Kitchen environment with randomized scene configurations to further test generalization. This requires collecting an entirely separate dataset, since the object positions varies on each episode. The environment was constructed by modifying the ``Kitchenshift'' domain~\cite{xing2021kitchenshift}. Both object types and their locations in the environment are randomized, which requires the agent to learn robust and general grasping strategies. There are several types of visual distractors, including randomized textures and lighting conditions. The observation space consists of three $128\times 128$ images: two side-view cameras and a wrist camera, as well as robot proprioception. The exact camera positions are also continuously randomized. Observations from different episodes are included in Fig. \ref{fig:domain_shift}. This level of variability introduces a significant challenge in terms of robustness and representation learning, reflecting challenges likely to be seen in the real world.

\textbf{Datasets}: To provide offline training data in this domain, we manually collected close to 20 hours of human teleoperation data:
\begin{enumerate}[leftmargin=*]

\item\textbf{Demonstrations:} We collected 500 expert-level demonstrations from an experienced teleoperator for the microwave, kettle, light switch and slide cabinet task (the same as the ``Mixed" dataset from Section \ref{sec:franka_standard}). This dataset is suitable for testing capabilities of representation learning approaches and benchmarking imitation learning algorithms. 

\item\textbf{Play:} We collect a datasets of 1000 episodes, which are not task-oriented from multiple operators with different levels of skills. The episodes consist of undirected environment interactions and involve manipulating between 2 to 6 objects in random order and placement. These episodes were collected by several tele-operators with different levels of experience, which introduces significant multi-modality in the data both in terms of behaviours and quality of executed grasps. 

\item\textbf{Sub-optimal Expert:} We also include a sub-optimal expert dataset consisting of 500 episodes, collected by inexperienced teleoperators, but we do not explicitly benchmark it in this work.
\end{enumerate}

\textbf{Tasks:} On the Demonstrations dataset, the agent is evaluated on the task corresponding to that demonstration. On the Play dataset, similar to Section \ref{sec:franka_standard}, we consider two tasks:

    \begin{enumerate}[leftmargin=*]
        \item \textbf{Mixed:} Similar to before in this task we need to manipulate the the microwave, kettle, light switch and slide cabinet objects. However, in addition to the representation learning an d robustness challenges that the randomized kitchen poses, the agent needs to learn from diverse data of varying quality. Another challenge is that while there are several episodes which manipulate all four objects, they do so in a different order, which creates a challenging problem for dynamic programming with multi-modal solutions.

        \item \textbf{Partial:} Similar to before, the agent needs to manipulate the microwave, kettle, bottom burner knob and light switch objects, which are never solved in the same episode in the offline data. 
    \end{enumerate}

\vspace{-0.3cm}
\subsection{Multi-Stage Manipulation with Scripted Data}
\vspace{-0.25cm}

\begin{wrapfigure}{r}{0.3\textwidth}
\vspace{-0.55cm}
  \begin{center}
    \includegraphics[width=0.29\textwidth]{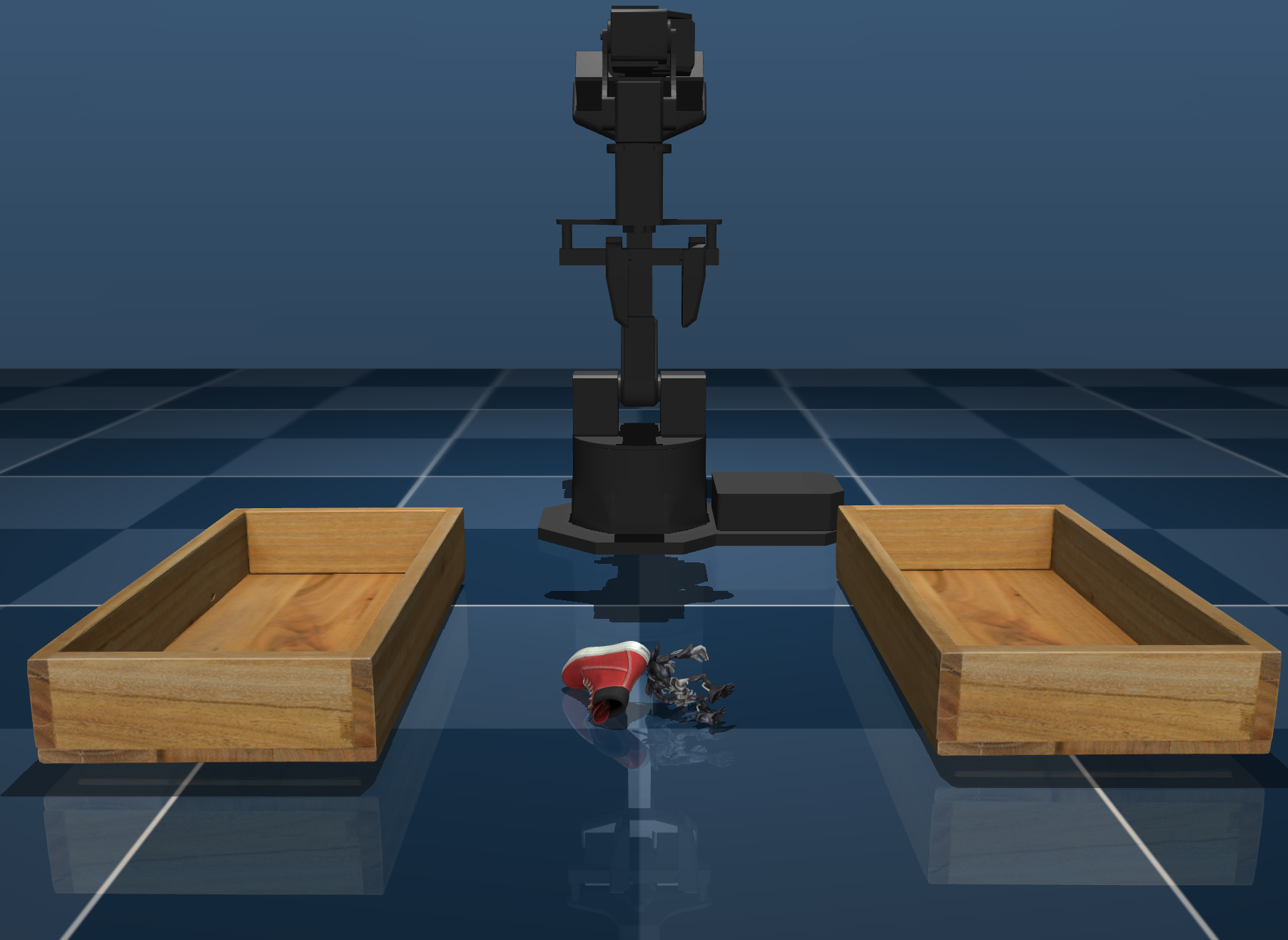}
  \end{center}
  \vspace{-0.35cm}
  \caption{\footnotesize{Setup for the Multi-Stage Manipulation with Scripted Data tasks consist of a simulated WidowX arm with 2 identical bins. In the center of the scene are two objects that are categorized as shoes or toys which the agent has to sort into their respective bins.}}
  \vspace{-0.35cm}
  \label{fig:sorting_setup}
\end{wrapfigure}

The goal of this task is to study composition of sub-optimal trajectories to solve longer-horizon tasks, incorporate visual observations, and handle data from weak scripted policies. These ingredients reflect problems that are often encountered in offline robotic RL, where we might want to compose longer-horizon behaviors out of datasets depicting individual primitive skills~\citep{fang2022planning,rosete2023latent,fang2023generalization}. To this end, we introduce a multi-stage bin sorting task. The simulated robot is a 6-DOF WidowX arm placed in front of two identical white bins with 2 objects to sort. These two objects are from two different categories: shoes and toys, and are taken from the Google Scanned Objects Dataset~\citep{downs2022google}, comprising 3D scans of real household objects.
 
\begin{figure}[t]
    \centering
    \includegraphics[width = \textwidth]{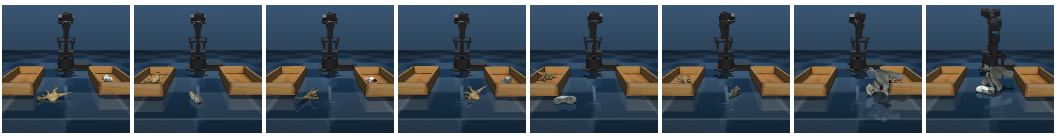}   
    \includegraphics[width = \textwidth]{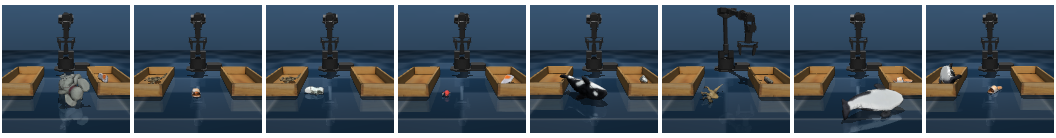}   
    \caption{\small Observations from the Multi-Stage Manipulation environments consist of a single $128\times 128$ RGB image from a side camera and robot proprioception. The environment includes several different types of shoes and toys, which require different grasps. Moreover, their locations are randomized across the scene. Textures, lighting conditions, and camera angles are also varied across episodes.}
    \label{fig:sorting_dataset}
\end{figure}

\textbf{Task:} As seen in Figure in Figure~\ref{fig:sorting_setup}, The objective is to sort each object into its respective bins. One bin corresponds to shoes and the other bin corresponds to toys. The reward function is the number of objects correctly sorted into each bin, where a "+1" reward is given when any of the objects are placed in their correct bins and a "+2" reward is given when both objects are sorted correctly. This task must be done from $128 \times 128 \times 3$ a combination of visual observations and the proprioceptive state of the robot (joint positions). There are multiple variations in which objects are seen. In each environment reset, one toy and one shoe will be randomly selected from a pool of 5 objects and placed in the central region of the scene. Various sample image observations can be seen in Figure~\ref{fig:sorting_dataset}.

\textbf{Datasets:} There are 2 tasks corresponding to the environments wx-sorting-v0 and wx-sorting-pickplacedata-v0.  Below, we provide a description of each.

 \begin{enumerate}[leftmargin=*]
     \item \textbf{Sorting:} The first dataset comprises of data collected with a scripted policy that attempts sorting both objects into their respective bins. The scripted policy with some likelihood places the object in the correct bin if grasped and otherwise in the incorrect bin. In all, there are 2000 episodes presented to the agent, which are mostly unsuccessful at solving the full task but consistently solve the individual segments of the task in separate episodes.

    \item \textbf{Sorting with Pickplace Data:} This dataset only comprises of transitions where the robot picks any object and places it in its respective bin. The data is similar to the dataset above in that there is a likelihood that the scripted agent places the object in the wrong bin. In all, there are 2000 episodes presented to the agent, solving the sorting task only partially.
    
 \end{enumerate}

\vspace{-0.3cm}
\section{Benchmark Results}
\vspace{-0.3cm}

\begin{table*}[t]
    \centering
    \resizebox{\textwidth}{!}{%
    \begin{tabular}{cc|c|c|c|c|c|c|c|c}
        \toprule
        \multirow{2}{*}{\textbf{Environment}} & \multirow{2}{*}{\textbf{Task}} & \multicolumn{7}{c}{\textbf{Method}} \\ \cmidrule{3-10} 
         & & \textbf{BC} & \textbf{IQL} &  \textbf{CQL} & \textbf{CalQL} \citep{nakamoto2023calql} & \textbf{TD3 + BC} & \textbf{RLPD} \citep{ball2023rlpd}& \textbf{DDPM + BC}& \textbf{IDQL} \\ 
         \midrule
        \multirow{3}{*}{\textbf{Standard Kitchen}} & Mixed & $0.461\pm0.124$& $0.457\pm0.129$& $0.0\pm0.0$& $0.0\pm0.0$& $0.003\pm0.003$& $0.0 \pm 0.0$& $0.253\pm0.082$& $0.020\pm0.008$ \\ \cmidrule{2-10}
         & Partial & $0.474\pm0.063$& $0.427\pm0.116$& $0.0\pm0.0$& $0.0\pm0.0$& $0.053\pm0.075$& $0.0\pm0.0$& $0.163\pm0.054$& $0.087\pm0.021$ \\ \midrule
        \multirow{3}{*}{\textbf{Randomized Kitchen}} & Demos & $0.144\pm0.010$ & $0.174\pm0.031$ & $0.023\pm0.032$ & $0.023\pm0.016$ & $0.052\pm0.033$ & $0.025\pm0.036$& $0.126\pm0.016$ & $0.033\pm0.011$  \\ \cmidrule{2-10}
         & Mixed & $0.057\pm0.019$ & $0.027\pm0.0$ & $0.005\pm0.002$ & $0.004\pm0.001$ & $0.057\pm0.026$ & $0.017\pm0.024$& $0.105\pm0.016$ & $0.009\pm0.004$ \\ \cmidrule{2-10}
         & Partial & $ 0.072\pm0.019 $ & $0.048\pm0.015$ & $0.003\pm0.005$ & $0.001\pm0.001$ & $0.023\pm0.007$ & $0.008\pm0.012$ & $0.044\pm0.010$ & $0.002\pm0.001$  \\ \midrule
        \multirow{3}{*}{\textbf{Locomotion}} & a1-walk-v0 & $1.006\pm0.015$ & $0.962\pm0.007$ & $0.068\pm0.112$ & $-0.171\pm0.033$ & $0.549\pm0.178$ & $0.032\pm0.007$ & - & -\\ \cmidrule{2-10}
         & a1-run-v0 & $0.684\pm0.026$ & $0.932\pm0.006$ & $-0.067\pm0.045$ & $-0.206\pm0.086$ & $0.002\pm0.021$ & $0.002\pm0.021$ &- &-\\ \cmidrule{2-10}
         & a1-hiking-v0 & $0.956\pm0.004$ & $0.935\pm0.003$ & $0.0\pm0.004$ & $-0.013\pm0.008$ & $0.003\pm0.001$ & $0.003\pm0.001$ &- & - \\ \midrule
        \multirow{3}{*}{\textbf{WidowX}} & wx-sorting-v0 & 0.152 $\pm$ 0.032 & 0.021 $\pm$ 0.016 & 0.0 $\pm$ 0.0 & 0.0 $\pm$ 0.0 & 0.016 $\pm$ 0.022 & - & 0.041 & 0.173 \\ \cmidrule{2-10}
         & wx-sorting-pickplacedata-v0 & 0.084 $\pm$ 0.048 & 0.0 $\pm$ 0.0 & 0.0 $\pm$ 0.0 & 0.0 $\pm$ 0.0 & 0.0 $\pm$ 0.0 & - & 0.081 & 0.25 \\ \bottomrule
    \end{tabular}
    }
    \vspace{2pt}
    \caption{\textbf{Evaluation of offline methods} for each task and dataset.}
    \label{tab:results_offline}
\end{table*}

\begin{table*}[t]
    \centering
    \resizebox{\textwidth}{!}{%
    \begin{tabular}{cc|c|c|c|c|c|c|c}
        \toprule
        \multirow{2}{*}{\textbf{Environment}} & \multirow{2}{*}{\textbf{Task}} & \multicolumn{6}{c}{\textbf{Method}} \\ \cmidrule{3-9} 
         & &  \textbf{IQL} &  \textbf{CQL} & \textbf{CalQL} \citep{nakamoto2023calql} & \textbf{TD3 + BC} & \textbf{RLPD} \citep{ball2023rlpd}& \textbf{DDPM + BC}& \textbf{IDQL} \\ 
         \midrule
        \multirow{3}{*}{\textbf{Standard Kitchen}} & Mixed & $0.123\pm0.102$& $0.0\pm0.0$& $0.0\pm0.0$& $0.067\pm0.066$& $0.139\pm0.075$ & $0.200\pm0.029$& $0.020\pm0.008$ \\ \cmidrule{2-9}
         & Partial  & $0.290\pm0.064$& $0.0\pm0.0$& $0.0\pm0.0$& $0.093\pm0.059$& $0.221\pm0.113$ & $0.177\pm0.022$& $0.087\pm0.021$ \\ \midrule
        \multirow{3}{*}{\textbf{Randomized Kitchen}} & Demos  & $0.234\pm0.017$ & $0.0\pm0.0$ & $0.023\pm0.016$ & $0.052\pm0.033$ & $0.001\pm0.001$ & $0.166\pm0.029$ & $0.033\pm0.011$  \\ \cmidrule{2-9}
         & Mixed & $0.25\pm0.0$ & $0.0\pm0.0$ & $0.004\pm0.001$ & $0.057\pm0.026$ & $0.011\pm0.009$  & $0.133\pm0.004$ & $0.009\pm0.004$ \\ \cmidrule{2-9}
         & Partial & $0.021\pm0.009$ & $0.0\pm0.0$ & $0.001\pm0.001$ & $0.023\pm0.007$ & $0.0\pm0.0$  & $0.084\pm0.009$ & $0.002\pm0.001$  \\ \midrule
        \multirow{3}{*}{\textbf{Locomotion}} & a1-walk-v0  & $0.935\pm0.017$ & $0.068\pm0.112$ & $0.750\pm0.027$ & $0.030\pm0.003$ & $1.016\pm0.005$ & - & - \\ \cmidrule{2-9}
         & a1-run-v0 & $0.936\pm0.021$ & $-0.067\pm0.045$ & $0.700\pm0.066$ & $0.110\pm0.091$ & $1.011\pm0.007$ &- & -\\ \cmidrule{2-9}
         & a1-hiking-v0 & $0.927\pm0.014$ & $0.0\pm0.004$ & $0.368\pm0.107$ & $0.938\pm0.015$ & $1.058\pm0.020$ & - & - \\
        \bottomrule
    \end{tabular}
    }
    \caption{\textbf{Evaluation of offline-to-online methods} for each task and dataset.}
    \label{tab:results_online}
\end{table*}

For each of the datasets in each of the domains, we evaluated a collection of recently proposed offline RL algorithms, as well as methods designed for online RL training with offline data (either via pre-training or joint training). We selected a range of algorithms that are meant to be representative of various different types of approaches. Although our evaluation algorithms do not cover every recent method (as there are many of them), we evaluated 8 separate algorithms, and we hope that in collaboration with the community, we can include many more evaluation numbers as part of the D5RL open-source repository.
We chose CQL~\citep{kumar2020conservative} as a standard representative example of a pessimistic/conservative offline RL method, together with Cal-QL~\citep{nakamoto2023calql}, a variant of CQL adapted for online finetuning. To evaluate implicit TD backups, we include IQL~\citep{kostrikov2021offline}, as well as IDQL~\citep{hansenestruch2023idql}, a recent extension of IQL that utilizes diffusion model policies. To evaluate BC-based regularization, we include TD3+BC~\citep{fujimoto2021minimalist}. We include RLPD~\citep{ball2023rlpd} as a representative example of a joint training method that runs online RL with prior data included in the buffer, and a behavioral cloning (BC) baseline as a diagnostic of the average performance in the dataset.

The results for all of the offline RL methods are included in Table~\ref{tab:results_offline}, with results after online finetuning included in Table~\ref{tab:results_online}. For completeness, we include RLPD in the offline results (using the same exact algorithm but without online collection). The online results are obtained by finetuning the offline value function and policy for each method, except for RLPD, where the online run is completely separate from the offline one. Further details about the specific training setup, hyperparameters, and number of update steps for each method are provided in Appendix B.

The results show that our proposed benchmark leaves considerable room for improvement for current offline RL and online finetuning methods. A few particularly prominent challenges include handling generalization and visual observations, and handling multi-stage tasks. When using image observations for the Franka kitchen tasks, particularly the more complex randomized domain, we see that many of the current RL methods struggle to exceed the performance of the simple behavioral cloning policy, indicating significant difficulties in learning robust perception. When learning the multi-stage WidowX tasks, we similarly see low performance, and in fact the na\"ive BC policy performs marginally better, again suggesting difficulties with scaling current RL methods into these domains. We believe that these results indicate that our benchmark provides significant room for improvement, and can drive development of more effective and scalable methods.



\vspace{-0.3cm}
\section{Discussion}
\vspace{-0.3cm}

We introduced a new benchmark for offline RL and online training with offline data, which we call D5RL. The aim of D5RL is to provide coverage of a variety of offline RL and online finetuning challenges, including different data compositions (scripted, human play-style data, and other sources), different input modalities (images and state), and tasks that require varying degrees of stitching, online finetuning, and generalization over initial state variability. Although the D5RL tasks are designed primarily for iterating on RL algorithms, all of the D5RL tasks are also designed to be reasonably reflective of real-world robotic tasks, with each environment containing a simulation of a real-world robot (an A1 quadruped, a Franka industrial arm, or a WidowX low-cost robotic arm) based on the robot's actual URDF, and tasks that reflect behaviors those robots might be expected to carry out in the real world. We also conducted an investigation with a number of existing offline RL and online finetuning methods to provide initial evaluation numbers with our benchmark, which we hope the community will utilize to develop more effective algorithms.

While we believe our benchmark provides a significant improvement over existing offline RL benchmark tasks, many of which are either saturated due to recent algorithm developments or do not cover as many of the problem dimensions as D5RL, our benchmark does have several limitations. First, we focus entirely on simulated robotics tasks. Such tasks are appealing because they cover complex dynamics and visual perception, but many aspects that make RL difficult in other domains, such as a high degree of stochasticity (e.g., in algorithmic trading) are absent in these domains. Benchmark tasks that address such domains would be very valuable and complementary to ours. Second, while our tasks reflect real-world robots, there is a limit to how realistic such simulated domains can be. Of course real data would be a ``gold standard'' in realism, but evaluating policies trained on real data would require either bridging the domain gap to simulation, or else using real physical robots, both of which would require considerable engineering and slow down the iteration cycle for algorithm developers. We therefore opted for a more conventional simulated evaluation to facilitate fast algorithms development, but we also believe that a real-world counterpart to D5RL would be valuable for the community. In conclusion, we hope that D5RL will serve as a new benchmark task for development of offline RL and online finetuning methods, and that future work can address some of the remaining blind spots of this benchmark to provide even comprehensive evaluations and facilitate more broadly applicable algorithms.

\bibliography{main}

\begin{thebibliography}{61}
\providecommand{\natexlab}[1]{#1}
\providecommand{\url}[1]{\texttt{#1}}
\expandafter\ifx\csname urlstyle\endcsname\relax
  \providecommand{\doi}[1]{doi: #1}\else
  \providecommand{\doi}{doi: \begingroup \urlstyle{rm}\Url}\fi

\bibitem[Agarwal et~al.(2020)Agarwal, Schuurmans, and Norouzi]{agarwal2020optimistic}
Rishabh Agarwal, Dale Schuurmans, and Mohammad Norouzi.
\newblock An optimistic perspective on offline reinforcement learning.
\newblock In \emph{International Conference on Machine Learning}, pp.\  104--114. PMLR, 2020.

\bibitem[Agarwal et~al.(2021)Agarwal, Schwarzer, Castro, Courville, and Bellemare]{agarwal2021deep}
Rishabh Agarwal, Max Schwarzer, Pablo~Samuel Castro, Aaron~C Courville, and Marc Bellemare.
\newblock Deep reinforcement learning at the edge of the statistical precipice.
\newblock \emph{Advances in neural information processing systems}, 34:\penalty0 29304--29320, 2021.

\bibitem[Ball et~al.(2023)Ball, Smith, Kostrikov, and Levine]{ball2023rlpd}
Philip~J. Ball, Laura Smith, Ilya Kostrikov, and Sergey Levine.
\newblock Efficient online reinforcement learning with offline data.
\newblock \emph{International Conference on Machine Learning (ICML)}, 2023.

\bibitem[Brockman et~al.(2016)Brockman, Cheung, Pettersson, Schneider, Schulman, Tang, and Zaremba]{brockman2016openai}
Greg Brockman, Vicki Cheung, Ludwig Pettersson, Jonas Schneider, John Schulman, Jie Tang, and Wojciech Zaremba.
\newblock Openai gym.
\newblock \emph{arXiv preprint arXiv:1606.01540}, 2016.

\bibitem[Cheng et~al.(2022)Cheng, Xie, Jiang, and Agarwal]{cheng2022adversarially}
Ching-An Cheng, Tengyang Xie, Nan Jiang, and Alekh Agarwal.
\newblock Adversarially trained actor critic for offline reinforcement learning.
\newblock In \emph{International Conference on Machine Learning}, pp.\  3852--3878. PMLR, 2022.

\bibitem[Cobbe et~al.(2019)Cobbe, Klimov, Hesse, Kim, and Schulman]{cobbe2019quantifying}
Karl Cobbe, Oleg Klimov, Chris Hesse, Taehoon Kim, and John Schulman.
\newblock Quantifying generalization in reinforcement learning.
\newblock In \emph{International Conference on Machine Learning}, pp.\  1282--1289. PMLR, 2019.

\bibitem[Cobbe et~al.(2020)Cobbe, Hesse, Hilton, and Schulman]{cobbe2020leveraging}
Karl Cobbe, Chris Hesse, Jacob Hilton, and John Schulman.
\newblock Leveraging procedural generation to benchmark reinforcement learning.
\newblock In \emph{International conference on machine learning}, pp.\  2048--2056. PMLR, 2020.

\bibitem[Contributors(2022)]{menagerie2022github}
MuJoCo~Menagerie Contributors.
\newblock {MuJoCo Menagerie: A collection of high-quality simulation models for MuJoCo}, 2022.
\newblock URL \url{http://github.com/deepmind/mujoco_menagerie}.

\bibitem[Dosovitskiy et~al.(2017)Dosovitskiy, Ros, Codevilla, Lopez, and Koltun1]{dosovitskiy2017carla}
Alexey Dosovitskiy, German Ros, Felipe Codevilla, Antonio Lopez, and Vladlen Koltun1.
\newblock Carla: An open urban driving simulator.
\newblock \emph{Conference on Robot Learning (CORL)}, 2017.

\bibitem[Downs et~al.(2022)Downs, Francis, Koenig, Kinman, Hickman, Reymann, McHugh, and Vanhoucke]{downs2022google}
Laura Downs, Anthony Francis, Nate Koenig, Brandon Kinman, Ryan Hickman, Krista Reymann, Thomas~B McHugh, and Vincent Vanhoucke.
\newblock Google scanned objects: A high-quality dataset of 3d scanned household items.
\newblock In \emph{2022 International Conference on Robotics and Automation (ICRA)}, pp.\  2553--2560. IEEE, 2022.

\bibitem[Duan et~al.(2016)Duan, Chen, Houthooft, Schulman, and Abbeel]{duan2016benchmarking}
Yan Duan, Xi~Chen, Rein Houthooft, John Schulman, and Pieter Abbeel.
\newblock Benchmarking deep reinforcement learning for continuous control.
\newblock In \emph{International conference on machine learning}, pp.\  1329--1338. PMLR, 2016.

\bibitem[Dulac-Arnold et~al.(2021)Dulac-Arnold, Levine, Mankowitz, Li, Paduraru, Gowal, and Hester]{dulac2021challenges}
Gabriel Dulac-Arnold, Nir Levine, Daniel~J Mankowitz, Jerry Li, Cosmin Paduraru, Sven Gowal, and Todd Hester.
\newblock Challenges of real-world reinforcement learning: definitions, benchmarks and analysis.
\newblock \emph{Machine Learning}, 110\penalty0 (9):\penalty0 2419--2468, 2021.

\bibitem[{Espeholt} et~al.(2018){Espeholt}, {Soyer}, {Munos}, {Simonyan}, {Mnih}, {Ward}, {Doron}, {Firoiu}, {Harley}, {Dunning}, {Legg}, and {Kavukcuoglu}]{2018arXiv180201561E}
Lasse {Espeholt}, Hubert {Soyer}, Remi {Munos}, Karen {Simonyan}, Volodymir {Mnih}, Tom {Ward}, Yotam {Doron}, Vlad {Firoiu}, Tim {Harley}, Iain {Dunning}, Shane {Legg}, and Koray {Kavukcuoglu}.
\newblock {IMPALA: Scalable Distributed Deep-RL with Importance Weighted Actor-Learner Architectures}.
\newblock \emph{arXiv e-prints}, art. arXiv:1802.01561, February 2018.
\newblock \doi{10.48550/arXiv.1802.01561}.

\bibitem[Fang et~al.(2022)Fang, Yin, Nair, and Levine]{fang2022planning}
Kuan Fang, Patrick Yin, Ashvin Nair, and Sergey Levine.
\newblock Planning to practice: Efficient online fine-tuning by composing goals in latent space.
\newblock In \emph{2022 IEEE/RSJ International Conference on Intelligent Robots and Systems (IROS)}, pp.\  4076--4083. IEEE, 2022.

\bibitem[Fang et~al.(2023)Fang, Yin, Nair, Walke, Yan, and Levine]{fang2023generalization}
Kuan Fang, Patrick Yin, Ashvin Nair, Homer~Rich Walke, Gengchen Yan, and Sergey Levine.
\newblock Generalization with lossy affordances: Leveraging broad offline data for learning visuomotor tasks.
\newblock In \emph{Conference on Robot Learning}, pp.\  106--117. PMLR, 2023.

\bibitem[Fu et~al.(2020)Fu, Kumar, Nachum, Tucker, and Levine]{fu2020d4rl}
Justin Fu, Aviral Kumar, Ofir Nachum, George Tucker, and Sergey Levine.
\newblock D4rl: Datasets for deep data-driven reinforcement learning.
\newblock \emph{arXiv preprint arXiv:2004.07219}, 2020.

\bibitem[Fujimoto \& Gu(2021)Fujimoto and Gu]{fujimoto2021minimalist}
Scott Fujimoto and Shixiang~Shane Gu.
\newblock A minimalist approach to offline reinforcement learning.
\newblock \emph{Advances in neural information processing systems}, 34:\penalty0 20132--20145, 2021.

\bibitem[Fujimoto et~al.(2019)Fujimoto, Meger, and Precup]{fujimoto2019off}
Scott Fujimoto, David Meger, and Doina Precup.
\newblock Off-policy deep reinforcement learning without exploration.
\newblock In \emph{International conference on machine learning}, pp.\  2052--2062. PMLR, 2019.

\bibitem[Gulcehre et~al.(2020)Gulcehre, Wang, Novikov, Paine, G{\'o}mez, Zolna, Agarwal, Merel, Mankowitz, Paduraru, et~al.]{gulcehre2020rl}
Caglar Gulcehre, Ziyu Wang, Alexander Novikov, Thomas Paine, Sergio G{\'o}mez, Konrad Zolna, Rishabh Agarwal, Josh~S Merel, Daniel~J Mankowitz, Cosmin Paduraru, et~al.
\newblock Rl unplugged: A suite of benchmarks for offline reinforcement learning.
\newblock \emph{Advances in Neural Information Processing Systems}, 33:\penalty0 7248--7259, 2020.

\bibitem[Gupta et~al.(2019{\natexlab{a}})Gupta, Kumar, Lynch, Levine, and Hausman]{gupta2019relay}
Abhishek Gupta, Vikash Kumar, Corey Lynch, Sergey Levine, and Karol Hausman.
\newblock Relay policy learning: Solving long-horizon tasks via imitation and reinforcement learning.
\newblock \emph{arXiv preprint arXiv:1910.11956}, 2019{\natexlab{a}}.

\bibitem[Gupta et~al.(2019{\natexlab{b}})Gupta, Kumar, Lynch, Levine, and Hausman]{lynch2019play}
Abhishek Gupta, Vikash Kumar, Corey Lynch, Sergey Levine, and Karol Hausman.
\newblock Relay policy learning: Solving long horizon tasks via imitation and reinforcement learning.
\newblock \emph{Conference on Robot Learning (CoRL)}, 2019{\natexlab{b}}.

\bibitem[Hansen-Estruch et~al.(2023)Hansen-Estruch, Kostrikov, Janner, Kuba, and Levine]{hansenestruch2023idql}
Philippe Hansen-Estruch, Ilya Kostrikov, Michael Janner, Jakub~Grudzien Kuba, and Sergey Levine.
\newblock Idql: Implicit q-learning as an actor-critic method with diffusion policies, 2023.

\bibitem[Henderson et~al.(2018)Henderson, Islam, Bachman, Pineau, Precup, and Meger]{henderson2018deep}
Peter Henderson, Riashat Islam, Philip Bachman, Joelle Pineau, Doina Precup, and David Meger.
\newblock Deep reinforcement learning that matters.
\newblock In \emph{Proceedings of the AAAI conference on artificial intelligence}, volume~32, 2018.

\bibitem[Hester et~al.(2018)Hester, Vecerik, Pietquin, Lanctot, Schaul, Piot, Horgan, Quan, Sendonaris, Osband, et~al.]{hester2018deep}
Todd Hester, Matej Vecerik, Olivier Pietquin, Marc Lanctot, Tom Schaul, Bilal Piot, Dan Horgan, John Quan, Andrew Sendonaris, Ian Osband, et~al.
\newblock Deep q-learning from demonstrations.
\newblock In \emph{Proceedings of the AAAI Conference on Artificial Intelligence}, volume~32, 2018.

\bibitem[Hsu et~al.(2022)Hsu, Kim, Rafailov, Wu, and Finn]{hsu2022seeing}
Kyle Hsu, Moo~Jin Kim, Rafael Rafailov, Jiajun Wu, and Chelsea Finn.
\newblock Vision-based manipulators need to also see from their hands.
\newblock \emph{International Conference on Learning Representations (ICLR)}, 2022.

\bibitem[Hubbs et~al.(2020)Hubbs, Perez, Sarwar, Sahinidis, Grossmann, and Wassick]{hubbs2020or}
Christian~D Hubbs, Hector~D Perez, Owais Sarwar, Nikolaos~V Sahinidis, Ignacio~E Grossmann, and John~M Wassick.
\newblock Or-gym: A reinforcement learning library for operations research problems.
\newblock \emph{arXiv preprint arXiv:2008.06319}, 2020.

\bibitem[Kalashnikov et~al.(2018)Kalashnikov, Irpan, Pastor, Ibarz, Herzog, Jang, Quillen, Holly, Kalakrishnan, Vanhoucke, et~al.]{kalashnikov2018qt}
Dmitry Kalashnikov, Alex Irpan, Peter Pastor, Julian Ibarz, Alexander Herzog, Eric Jang, Deirdre Quillen, Ethan Holly, Mrinal Kalakrishnan, Vincent Vanhoucke, et~al.
\newblock Qt-opt: Scalable deep reinforcement learning for vision-based robotic manipulation.
\newblock \emph{arXiv preprint arXiv:1806.10293}, 2018.

\bibitem[Kostrikov et~al.(2020)Kostrikov, Yarats, and Fergus]{kostrikov2020image}
Ilya Kostrikov, Denis Yarats, and Rob Fergus.
\newblock Image augmentation is all you need: Regularizing deep reinforcement learning from pixels.
\newblock \emph{arXiv preprint arXiv:2004.13649}, 2020.

\bibitem[Kostrikov et~al.(2021)Kostrikov, Nair, and Levine]{kostrikov2021offline}
Ilya Kostrikov, Ashvin Nair, and Sergey Levine.
\newblock Offline reinforcement learning with implicit q-learning.
\newblock \emph{arXiv preprint arXiv:2110.06169}, 2021.

\bibitem[Krizhevsky et~al.(2017)Krizhevsky, Sutskever, and Hinton]{krizhevsky2017imagenet}
Alex Krizhevsky, Ilya Sutskever, and Geoffrey~E Hinton.
\newblock Imagenet classification with deep convolutional neural networks.
\newblock \emph{Communications of the ACM}, 60\penalty0 (6):\penalty0 84--90, 2017.

\bibitem[Kumar et~al.(2019)Kumar, Fu, Soh, Tucker, and Levine]{kumar2019stabilizing}
Aviral Kumar, Justin Fu, Matthew Soh, George Tucker, and Sergey Levine.
\newblock Stabilizing off-policy q-learning via bootstrapping error reduction.
\newblock \emph{Advances in Neural Information Processing Systems}, 32, 2019.

\bibitem[Kumar et~al.(2020)Kumar, Zhou, Tucker, and Levine]{kumar2020conservative}
Aviral Kumar, Aurick Zhou, George Tucker, and Sergey Levine.
\newblock Conservative q-learning for offline reinforcement learning.
\newblock \emph{Advances in Neural Information Processing Systems}, 33:\penalty0 1179--1191, 2020.

\bibitem[Kuo et~al.(2022)Kuo, Polizzotto, Finfer, Garcia, S{\"o}nnerborg, Zazzi, B{\"o}hm, Kaiser, Jorm, and Barbieri]{kuo2022health}
Nicholas I-Hsien Kuo, Mark~N Polizzotto, Simon Finfer, Federico Garcia, Anders S{\"o}nnerborg, Maurizio Zazzi, Michael B{\"o}hm, Rolf Kaiser, Louisa Jorm, and Sebastiano Barbieri.
\newblock The health gym: synthetic health-related datasets for the development of reinforcement learning algorithms.
\newblock \emph{Scientific Data}, 9\penalty0 (1):\penalty0 693, 2022.

\bibitem[Kurenkov \& Kolesnikov(2022)Kurenkov and Kolesnikov]{kurenkov2022showing}
Vladislav Kurenkov and Sergey Kolesnikov.
\newblock Showing your offline reinforcement learning work: Online evaluation budget matters.
\newblock In \emph{International Conference on Machine Learning}, pp.\  11729--11752. PMLR, 2022.

\bibitem[LeCun et~al.(2015)LeCun, Bengio, and Hinton]{lecun2015deep}
Yann LeCun, Yoshua Bengio, and Geoffrey Hinton.
\newblock Deep learning.
\newblock \emph{nature}, 521\penalty0 (7553):\penalty0 436--444, 2015.

\bibitem[Levine et~al.(2020)Levine, Kumar, Tucker, and Fu]{levine2020offline}
Sergey Levine, Aviral Kumar, George Tucker, and Justin Fu.
\newblock Offline reinforcement learning: Tutorial, review, and perspectives on open problems.
\newblock \emph{arXiv preprint arXiv:2005.01643}, 2020.

\bibitem[Liu et~al.(2022)Liu, Xia, Rui, Gao, Yang, Zhu, Wang, Wang, and Guo]{liu2022finrl}
Xiao-Yang Liu, Ziyi Xia, Jingyang Rui, Jiechao Gao, Hongyang Yang, Ming Zhu, Christina Wang, Zhaoran Wang, and Jian Guo.
\newblock Finrl-meta: Market environments and benchmarks for data-driven financial reinforcement learning.
\newblock \emph{Advances in Neural Information Processing Systems}, 35:\penalty0 1835--1849, 2022.

\bibitem[Lu et~al.(2022)Lu, Ball, Rudner, Parker-Holder, Osborne, and Teh]{lu2022challenges}
Cong Lu, Philip~J Ball, Tim~GJ Rudner, Jack Parker-Holder, Michael~A Osborne, and Yee~Whye Teh.
\newblock Challenges and opportunities in offline reinforcement learning from visual observations.
\newblock \emph{arXiv preprint arXiv:2206.04779}, 2022.

\bibitem[Lynch et~al.(2020)Lynch, Khansari, Xiao, Kumar, Tompson, Levine, and Sermanet]{lynch2020learning}
Corey Lynch, Mohi Khansari, Ted Xiao, Vikash Kumar, Jonathan Tompson, Sergey Levine, and Pierre Sermanet.
\newblock Learning latent plans from play.
\newblock In \emph{Conference on robot learning}, pp.\  1113--1132. PMLR, 2020.

\bibitem[Mandlekar et~al.(2021)Mandlekar, Xu, Wong, Nasiriany, Wang, Kulkarni, Fei-Fei, Savarese, Zhu, and Mart{\'\i}n-Mart{\'\i}n]{mandlekar2021matters}
Ajay Mandlekar, Danfei Xu, Josiah Wong, Soroush Nasiriany, Chen Wang, Rohun Kulkarni, Li~Fei-Fei, Silvio Savarese, Yuke Zhu, and Roberto Mart{\'\i}n-Mart{\'\i}n.
\newblock What matters in learning from offline human demonstrations for robot manipulation.
\newblock \emph{arXiv preprint arXiv:2108.03298}, 2021.

\bibitem[Mendonca et~al.(2021)Mendonca, Rybkin, Daniilidis, Hafner, and Pathak]{lexa2021}
Russell Mendonca, Oleh Rybkin, Kostas Daniilidis, Danijar Hafner, and Deepak Pathak.
\newblock Discovering and achieving goals via world models, 2021.

\bibitem[Mnih et~al.(2013)Mnih, Kavukcuoglu, Silver, Graves, Antonoglou, Wierstra, and Riedmiller]{mnih2013playing}
Volodymyr Mnih, Koray Kavukcuoglu, David Silver, Alex Graves, Ioannis Antonoglou, Daan Wierstra, and Martin Riedmiller.
\newblock Playing atari with deep reinforcement learning.
\newblock \emph{arXiv preprint arXiv:1312.5602}, 2013.

\bibitem[Nair et~al.(2020)Nair, Gupta, Dalal, and Levine]{nair2020awac}
Ashvin Nair, Abhishek Gupta, Murtaza Dalal, and Sergey Levine.
\newblock Awac: Accelerating online reinforcement learning with offline datasets.
\newblock \emph{arXiv preprint arXiv:2006.09359}, 2020.

\bibitem[Nakamoto et~al.(2023)Nakamoto, Zhai, Singh, Mark, Ma, Finn, Kumar, and Levine]{nakamoto2023calql}
Mitsuhiko Nakamoto, Yuexiang Zhai, Anika Singh, Max~Sobol Mark, Yi~Ma, Chelsea Finn, Aviral Kumar, and Sergey Levine.
\newblock Cal-ql: Calibrated offline rl pre-training for efficient online fine-tuning.
\newblock \emph{ArXiv}, abs/2303.05479, 2023.

\bibitem[Nichol \& Dhariwal(2021)Nichol and Dhariwal]{nichol2021improved}
Alexander~Quinn Nichol and Prafulla Dhariwal.
\newblock Improved denoising diffusion probabilistic models.
\newblock In \emph{International Conference on Machine Learning}, pp.\  8162--8171. PMLR, 2021.

\bibitem[Osband et~al.(2019)Osband, Doron, Hessel, Aslanides, Sezener, Saraiva, McKinney, Lattimore, Szepesvari, Singh, et~al.]{osband2019behaviour}
Ian Osband, Yotam Doron, Matteo Hessel, John Aslanides, Eren Sezener, Andre Saraiva, Katrina McKinney, Tor Lattimore, Csaba Szepesvari, Satinder Singh, et~al.
\newblock Behaviour suite for reinforcement learning.
\newblock \emph{arXiv preprint arXiv:1908.03568}, 2019.

\bibitem[Qin et~al.(2022)Qin, Zhang, Gao, Chen, Li, Zhang, and Yu]{qin2022neorl}
Rong-Jun Qin, Xingyuan Zhang, Songyi Gao, Xiong-Hui Chen, Zewen Li, Weinan Zhang, and Yang Yu.
\newblock Neorl: A near real-world benchmark for offline reinforcement learning.
\newblock \emph{Advances in Neural Information Processing Systems}, 35:\penalty0 24753--24765, 2022.

\bibitem[Rosete-Beas et~al.(2023)Rosete-Beas, Mees, Kalweit, Boedecker, and Burgard]{rosete2023latent}
Erick Rosete-Beas, Oier Mees, Gabriel Kalweit, Joschka Boedecker, and Wolfram Burgard.
\newblock Latent plans for task-agnostic offline reinforcement learning.
\newblock In \emph{Conference on Robot Learning}, pp.\  1838--1849. PMLR, 2023.

\bibitem[{Singh} et~al.(2020){Singh}, {Yu}, {Yang}, {Zhang}, {Kumar}, and {Levine}]{2020arXiv201014500S}
Avi {Singh}, Albert {Yu}, Jonathan {Yang}, Jesse {Zhang}, Aviral {Kumar}, and Sergey {Levine}.
\newblock {COG: Connecting New Skills to Past Experience with Offline Reinforcement Learning}.
\newblock \emph{arXiv e-prints}, art. arXiv:2010.14500, October 2020.
\newblock \doi{10.48550/arXiv.2010.14500}.

\bibitem[Smith et~al.(2022)Smith, Kostrikov, and Levine]{Smith2022AWI}
Laura Smith, Ilya Kostrikov, and Sergey Levine.
\newblock A walk in the park: Learning to walk in 20 minutes with model-free reinforcement learning.
\newblock \emph{ArXiv}, abs/2208.07860, 2022.

\bibitem[Song et~al.(2022)Song, Zhou, Sekhari, Bagnell, Krishnamurthy, and Sun]{song2022hybrid}
Yuda Song, Yifei Zhou, Ayush Sekhari, J~Andrew Bagnell, Akshay Krishnamurthy, and Wen Sun.
\newblock Hybrid rl: Using both offline and online data can make rl efficient.
\newblock \emph{arXiv preprint arXiv:2210.06718}, 2022.

\bibitem[Szot et~al.(2021)Szot, Clegg, Undersander, Wijmans, Zhao, Turner, Maestre, Mukadam, Chaplot, Maksymets, et~al.]{szot2021habitat}
Andrew Szot, Alexander Clegg, Eric Undersander, Erik Wijmans, Yili Zhao, John Turner, Noah Maestre, Mustafa Mukadam, Devendra~Singh Chaplot, Oleksandr Maksymets, et~al.
\newblock Habitat 2.0: Training home assistants to rearrange their habitat.
\newblock \emph{Advances in Neural Information Processing Systems}, 34:\penalty0 251--266, 2021.

\bibitem[Todorov et~al.(2012)Todorov, Erez, and Tassa]{todorov2012mujoco}
Emanuel Todorov, Tom Erez, and Yuval Tassa.
\newblock Mujoco: A physics engine for model-based control.
\newblock In \emph{2012 IEEE/RSJ International Conference on Intelligent Robots and Systems}, pp.\  5026--5033. IEEE, 2012.
\newblock \doi{10.1109/IROS.2012.6386109}.

\bibitem[Tunyasuvunakool et~al.(2020)Tunyasuvunakool, Muldal, Doron, Liu, Bohez, Merel, Erez, Lillicrap, Heess, and Tassa]{tunyasuvunakool2020dm_control}
Saran Tunyasuvunakool, Alistair Muldal, Yotam Doron, Siqi Liu, Steven Bohez, Josh Merel, Tom Erez, Timothy Lillicrap, Nicolas Heess, and Yuval Tassa.
\newblock {dm\_control}: Software and tasks for continuous control.
\newblock \emph{Software Impacts}, 6:\penalty0 100022, 2020.

\bibitem[Vecerik et~al.(2017)Vecerik, Hester, Scholz, Wang, Pietquin, Piot, Heess, Roth{\"o}rl, Lampe, and Riedmiller]{vecerik2017leveraging}
Mel Vecerik, Todd Hester, Jonathan Scholz, Fumin Wang, Olivier Pietquin, Bilal Piot, Nicolas Heess, Thomas Roth{\"o}rl, Thomas Lampe, and Martin Riedmiller.
\newblock Leveraging demonstrations for deep reinforcement learning on robotics problems with sparse rewards.
\newblock \emph{arXiv preprint arXiv:1707.08817}, 2017.

\bibitem[Wang et~al.(2019)Wang, Bao, Clavera, Hoang, Wen, Langlois, Zhang, Zhang, Abbeel, and Ba]{wang2019benchmarking}
Tingwu Wang, Xuchan Bao, Ignasi Clavera, Jerrick Hoang, Yeming Wen, Eric Langlois, Shunshi Zhang, Guodong Zhang, Pieter Abbeel, and Jimmy Ba.
\newblock Benchmarking model-based reinforcement learning.
\newblock \emph{arXiv preprint arXiv:1907.02057}, 2019.

\bibitem[Wei et~al.(2022)Wei, Tay, Bommasani, Raffel, Zoph, Borgeaud, Yogatama, Bosma, Zhou, Metzler, et~al.]{wei2022emergent}
Jason Wei, Yi~Tay, Rishi Bommasani, Colin Raffel, Barret Zoph, Sebastian Borgeaud, Dani Yogatama, Maarten Bosma, Denny Zhou, Donald Metzler, et~al.
\newblock Emergent abilities of large language models.
\newblock \emph{arXiv preprint arXiv:2206.07682}, 2022.

\bibitem[Wu et~al.(2017)Wu, Kreidieh, Parvate, Vinitsky, and Bayen]{wu2017flow}
Cathy Wu, Aboudy Kreidieh, Kanaad Parvate, Eugene Vinitsky, and Alexandre~M Bayen.
\newblock Flow: Architecture and benchmarking for reinforcement learning in traffic control.
\newblock \emph{arXiv preprint arXiv:1710.05465}, 10, 2017.

\bibitem[Xia et~al.(2019)Xia, Li, Chen, Shen, Mart{\'\i}n-Mart{\'\i}n, Hirose, Zamir, Fei-Fei, and Savarese]{xia2019gibson}
Fei Xia, Chengshu Li, Kevin Chen, William~B Shen, Roberto Mart{\'\i}n-Mart{\'\i}n, Noriaki Hirose, Amir~R Zamir, Li~Fei-Fei, and Silvio Savarese.
\newblock Gibson env v2: Embodied simulation environments for interactive navigation.
\newblock \emph{Stanford University, Tech. Rep.}, 2019.

\bibitem[Xing et~al.(2021)Xing, Gupta, Powers*, and Dean*]{xing2021kitchenshift}
Eliot Xing, Abhinav Gupta, Sam Powers*, and Victoria Dean*.
\newblock Kitchenshift: Evaluating zero-shot generalization of imitation-based policy learning under domain shifts.
\newblock In \emph{NeurIPS 2021 Workshop on Distribution Shifts: Connecting Methods and Applications}, 2021.
\newblock URL \url{https://openreview.net/forum?id=DdglKo8hBq0}.

\bibitem[Yu et~al.(2020)Yu, Quillen, He, Julian, Hausman, Finn, and Levine]{yu2020meta}
Tianhe Yu, Deirdre Quillen, Zhanpeng He, Ryan Julian, Karol Hausman, Chelsea Finn, and Sergey Levine.
\newblock Meta-world: A benchmark and evaluation for multi-task and meta reinforcement learning.
\newblock In \emph{Conference on robot learning}, pp.\  1094--1100. PMLR, 2020.

\end{thebibliography}
\bibliographystyle{rlc}

\newpage
\appendix
\section{Environments}
\label{app:environments}

\subsection{Legged Locomotion}
We construct the locomotion tasks using MuJoCo~\cite{todorov2012mujoco} and DeepMind's dm\_control~\cite{tunyasuvunakool2020dm_control} suite, using the model of Unitree's A1 quadruped from MuJoCo Menagerie~\cite{menagerie2022github}. The robot only receives as input proprioceptive and goal information. In particular, the robot's observations consist of its root's local forward linear velocity, orientation (roll and pitch), angular velocity (roll, pitch, and yaw), and its (12) joint angles and velocities. We also append the previous action applied. For the hiking task, we include the displacement vector between the robot to the next way point along the hiking path. The reward function is a simple locomotion reward that encourages a particular velocity to be tracked, subject to penalties on the body's angular velocity. For exact details on the reward function, we refer to Section IV.B of~\citet{Smith2022AWI}. The robot's actions are PD targets for the 12 joints. 

\subsection{Standard Franka Kitchen Manipulation Environment}
\label{appendix:standard_kitchen}

For the Standard Franka Kitchen Manipulation environment, we make some slight modifications to the Franka Kitchen environment from \citet{gupta2019relay} (RPL). 
The RPL Franka Kitchen environment requires controlling a simulated 9-DOF Franka Emika Robot to manipulate a set of four pre-defined objects into a desired configuration. At each timestep, a reward of 1.0 is given for each object that is in the correct configuration, with the maximum reward possible at each timestep being 4.0. The action space is joint-space control commands to the robot. 

We modify the original camera angle of the RPL environment to be the camera angle used in the LEXA benchmark~\citep{lexa2021}. Additionally, we add a wrist camera. We render both cameras at 128x128 resolutions. The observation space consists of two RGB images from the two cameras concatenated together, plus robot proprioception.   

We also utilize frame stacking in our experiments. This amounts to stacking the previous three images along the channel dimension, allowing the agent to have a short history of observations from which it can estimate movement and velocity, as done in \citep{mnih2013playing}.

\subsection{Randomized Franka Kitchen Manipulation Environment}

The Randomized Franka Kitchen Environment modifies the ``Kitchenshift'' domain~\cite{xing2021kitchenshift}, which is itself a heavily modified version of the RPL Kitchen environment. The Randomized Kitchen environment includes a large degree of domain randomization and visual diversity. At the start of each episode, the initial positions of the objects are randomized, as well as textures and lighting conditions. The specific types of objects are randomized too (eg: one type of kettle can be switched for a differently shaped type of kettle). 
The underlying tasks and rewards are the same as in the Standard Kitchen Environment. The action space is the same as in the Standard Kitchen environment. 

We use three RGB cameras (two side cameras and one wrist camera), each rendered at a resolution of 128x128 pixels. Robot proprioception is also included in the observation space. Similar to the Standard Kitchen environment, we use a frame-stacking wrapper around the Randomized Kitchen environment to maintain a history of 3 images. 

\subsection{Multi-Stage Manipulation with Scripted Data}

The multi-stage bin sorting task is an environment constructed using the DeepMind's dm\_control, a software stack utilized for physics-based Simulation and RL environments. The WidowX 250 was specified with an XML file which includes information about the robot's joints with respect to their sizes and weight. A position-based controller was used for the robot, where a specified action was indicated as a change in robot position. This controller was a PID-based controller. The objects and containers were sourced from Google's Scanned Object Dataset~\citep{downs2022google}, which contains photo-realistic 3D object models. From here, we selected 2 identical bins as containers and a set of objects that lie in two categories: toys and shoes. The objects were scaled to be graspable by the robot and fit in the container and are placed in the scene in any orientation (random quaternion). The background was a static tabletop where the robot, containers, and objects were all placed as seen in Figure~\ref{fig:sorting_setup}. 
\newpage
\section{Datasets}
We summarize the datasets, their construction and composition, for each of the tasks (organized by environment).

\subsection{Legged Locomotion}
We trained 3 A1s with the goal of tracking 3 speeds: 0.5, 0.8, and 1.0 m/s using RL (with the same inputs and reward function as described in~\autoref{app:environments}). We then consolidated their replay buffers and relabeled them as if their goal was to track speeds of 0.75m/s and 1.25m/s for the Interpolate Speed and Extrapolate Speed tasks, respectively. For the Hiking task, we trained a direction-conditioned policy using RL, again with the same observation and action space.

\subsection{Standard Franka Kitchen Manipulation Environment}

For our experiments we re-render the original RPL datasets \citep{gupta2019relay} with the two cameras described in section \ref{appendix:standard_kitchen}. We add in proprioception to the observations, which consists of the 9 joint angles of the robot arm. The dataset contains $563$ trajectories, with $128,569$ total transitions. The average undiscounted episode return is $261.12$, and the average number of objects manipulated per episode is $3.98$.

\subsection{Randomized Franka Kitchen Manipulation Environment}

We collected three distinct datasets for the Randomized Kitchen environment using tele-operation: Demonstrations, Play, and Sub-Optimal Expert. The differences between these datasets are described in Section \ref{subsection:randomized_kitchen}. The Demonstrations dataset contains $500$ total trajectories, with $250,500$ total transitions. The average undiscounted episode return is $1148.78$, and the average number of objects manipulated per episode is $4.0$. The Play dataset contains $1,000$ total trajectories, with $501,000$ total transitions. The average undiscounted episode return is $870.50$, and the average number of objects manipulated per episode is $3.62$. The Sub-Optimal Expert dataset contains $500$ total trajectories, with $250,500$ total transitions. The average undiscounted episode return is $911.70$, and the average number of objects manipulated per episode is $3.55$.

\subsection{Multi-Stage Manipulation with Scripted Data}
For the Multi-Stage Bin Sorting Task, we used hand-engineered scripted policies. These scripted policies used information such as the position of the object as well as containers to solve their respective tasks of interest. These scripted policies were given a time horizon of 500 to solve this task. For the \textbf{pick and place} task, the scripted policies were constructed to randomly select one of the two objects in the scene to grasp. 70\% of the time, the policy moved the object toward the correct bin. Other times, the object was directed to the incorrect bin (mimicking a scenario that the object was misclassified and sorted into the incorrect bin). For the \textbf{sorting} task, the scripted policy completed both stages of the task by picking and placing each object in succession. Our scripted policies are inspired by the procedure used in COG~\citep{2020arXiv201014500S}. 
\newpage
\section{Baselines}

\subsection{Architecture Design Choice}
For tasks that require learning from visual observations, we utilize the Impala architecture for our experiments. For the actor and critic, the network backbone we used the architecture is found in Impala~\citep{2018arXiv180201561E}. For environments that relied on multiple camera viewpoints such as the Franka Kitchen environments, image observations were frame-stacked and passed into the network. The output of the neural network was flattened and passed through an MLP to construct the actor and critic networks for each method. For environments with a proprioceptive state, this observation was concatenated to the flattened output of the network, prior to being passed through the MLP network. 

\subsection{Methods + Implementation Details}

\begin{wraptable}{r}{5.5cm}
\vspace{-0.8cm}
\centering
\scalebox{0.8}{
\begin{tabular}{c c}
  \toprule
    \textbf{Hyperparameters} & \textbf{IQL}\\         
  \midrule
    $\tau$ & 0.5, 0.7, 0.9, 0.95  \\
   actor architecture &  Impala\\
   critic (Q/V) architecture &  Impala\\
   actor learning rate &  1e-4 \\
   critic (Q/V) learning rate &  3e-4  \\
    batch size &  64\\
  \bottomrule
\end{tabular} }
\caption{\footnotesize \textbf{Hyperparameters for IQL}. We primarily utilize the default hyperparameters as prescribed in the paper and sweep over the expectile $\tau$.}
\label{table:hyperparam_iql} 
\end{wraptable}
Here we describe the prior methods we evaluate and describe task-specific implementation details.

\textbf{IQL~\citep{kostrikov2021offline}}
For the implementation of IQL, we modify the open source implementation of IQL found in \url{https://github.com/ikostrikov/jaxrl2}. The hyperparameters utilized for both methods can be found in Table~\ref{table:hyperparam_iql}. During training, we utilize the data augmentations of color jitter and random crops as proposed in DrQ~\citep{kostrikov2020image} which allows for better generalization.

\begin{wraptable}{r}{5.5cm}
\centering
\scalebox{0.8}{
\begin{tabular}{c c}
  \toprule
    \textbf{Hyperparameters} & \textbf{CQL} and \textbf{CalQL}\\         
  \midrule
    $\alpha$ (online + offline) & 0.1, 1, 5, 10  \\
   actor architecture &  Impala\\
   critic architecture &  Impala\\
   actor learning rate &  1e-4 \\
   critic learning rate &  3e-4  \\
    batch size &  64\\
  \bottomrule
\end{tabular} }
\vspace{0.05cm}
\caption{\footnotesize \textbf{Hyperparameters for CQL and CalQL}. We primarily utilize the default hyperparameters as prescribed in the paper and sweep over the constant $\alpha$.}
\label{table:hyperparam_cql} 
\end{wraptable}
\textbf{CQL~\citep{kumar2020conservative}} and \textbf{CalQL~\citep{nakamoto2023calql}}
For the implementation, we modify the open source implementation of CQL found in \url{https://github.com/ikostrikov/jaxrl2} for CQL and CalQL. The hyperparameters utilized for both methods can be found in Table~\ref{table:hyperparam_cql}. During training, we utilize the data augmentations of color jitter and random crops as proposed in DrQ~\citep{kostrikov2020image} which allows for better generalization. For CalQL, the lower bound was computed with the Monte-Carlo returns calculated using the rewards of the collected demonstrations, following the recipe in \citet{nakamoto2023calql}. 

\textbf{TD3 + BC~\citep{fujimoto2019off}} 
TD3+BC is an offline RL method that modifies an online RL method for the offline regime by simply adding a BC term to encourage the policy to resemble the behavior policy. For pixel-based experiments, we use the open-source implementation found in \url{https://github.com/ikostrikov/jaxrl2}. For state-based experiments, we use the authors' implementation at: \url{https://github.com/sfujim/TD3_BC}.

\textbf{RLPD~\citep{ball2023rlpd}}
RLPD is a method for online RL with access to offline data that has demonstrated state-of-the-art results on tasks designed to evaluate fine-tuning from offline RL pre-training; therefore, we include it as a main baseline for the fine-tuning regime. For `fine-tuning' evaluation, we evaluate RLPD as designed, i.e., without pre-training. For offline evaluation, we adapt RLPD to only sample from the offline data. We use the implementation by the authors open-sourced at: \url{https://github.com/ikostrikov/rlpd} and use the default hyperparameters as prescribed in the paper for all environments. For the Standard Kitchen and Randomized Kitchen environments, we used an Impala network architecture for the policy and critic network encoders. 

\textbf{DDPM + BC}
For the implementation of DDPM+BC, we modify the implementation of the behavior cloning policy from IDQL~\citep{hansenestruch2023idql} from \url{https://github.com/philippe-eecs/IDQL}. This involves attaching the convolutional encoder used to the architecture proposed in IDQL (LayerNorm + ResNet). During training, we use data augmentations as proposed in DrQ~\citep{kostrikov2020image} which improve generalization.

Furthermore, instead of using the variance preserving schedule as used in IDQL, we use the cosine schedule~\citep{nichol2021improved} and $T=20$. All other hyperparameters for the diffusion process and trunk architecture are the same as in IDQL. While we train for 2 gradient million steps, we recommend training for longer as the diffusion objective takes longer to train.

\textbf{IDQL~\citep{hansenestruch2023idql}}
The IDQL implementation combines the IQL implementation with the DDPM+BC implementation. After training the $Q$-function using Pixel IQL and the diffusion behavior policy using DDPM+BC, we combine the two during inference and sample the diffusion policy $N$ times and select the action that receives the highest $Q$-value. We use $N=64$ as from IDQL, but we recommend tuning this hyperparameter as to avoid potential OOD samples. 


\end{document}